%% file: main.tex
\documentclass[conference]{IEEEtran}
\IEEEoverridecommandlockouts
\usepackage{cite}
\usepackage{amsmath,amssymb,amsfonts}
\usepackage{graphicx}
\usepackage{textcomp}
\usepackage{xcolor}
\usepackage{adjustbox}
\usepackage{multirow}
\usepackage{tabularx}
\usepackage{algorithm}
\usepackage{amsmath}
\usepackage{mathtools}
\usepackage{amssymb}
\usepackage{breqn}
\usepackage{booktabs}

\usepackage{caption}
\usepackage{subcaption}
\usepackage{algpseudocode}
\DeclareMathOperator{\E}{\mathbb{E}}

\DeclareMathOperator{\R}{\mathbb{R}}

\def\BibTeX{{\rm B\kern-.05em{\sc i\kern-.025em b}\kern-.08em
    T\kern-.1667em\lower.7ex\hbox{E}\kern-.125emX}}
    
\begin{document}

\title{Distill Knowledge in Multi-task Reinforcement Learning with Optimal-Transport Regularization}


\makeatletter
\newcommand{\linebreakand}{%
  \end{@IEEEauthorhalign}
  \hfill\mbox{}\par
  \mbox{}\hfill\begin{@IEEEauthorhalign}
}
\makeatother

\author{
\IEEEauthorblockN{Bang Giang Le}
\IEEEauthorblockA{\textit{HMI Lab} \\
\textit{VNU University of Engineering and Technology}\\
Hanoi, Vietnam}\\[0.1cm]
\and
\IEEEauthorblockN{Viet Cuong Ta}
\IEEEauthorblockA{\textit{HMI Lab} \\
\textit{VNU University of Engineering and Technology}\\
Hanoi, Vietnam}\\[0.1cm]
}

\maketitle

\begin{abstract}
\input{abstract}
\end{abstract}


\input{introduction}

\input{related_work}

\input{background}

\input{method}
\input{result}

\section{Conclusion}
In this work, we have proposed a new method for distilling knowledge between multi-tasks RL by augmenting task-specific rewards with an additional optimal transport-based objective.
The optimal transport is used to constrain the deviation one task can deviate from other tasks, which implicitly forms a mechanism for regularizing and sharing between agents.
We empirically demonstrate our proposed approach on three grid world-based environments.
Compared to the baselines, our sharing algorithm can help accelerate the training of multi agents in grid-based environments. Finally, we examine the effect of the OT rewards in each environment.


\section*{Acknowledgements}
This material is based upon work supported by the Air Force Office of Scientific Research under award number FA2386-22-1-4026.

\bibliography{ref.bib}
\bibliographystyle{IEEEtran}

\end{document}

%% file: abstract.tex
In multi-task reinforcement learning, it is possible to improve the data efficiency of training agents by transferring knowledge from other different but related tasks.
Because the experiences from different tasks are usually biased toward the specific task goals.
Traditional methods rely on Kullback-Leibler regularization to stabilize the transfer of knowledge from one task to the others.
In this work, we explore the direction of replacing the Kullback-Leibler divergence with a novel Optimal transport-based regularization.
By using the Sinkhorn mapping, we can approximate the Optimal transport distance between the state distribution of tasks.
The distance is then used as an amortized reward to regularize the amount of sharing information.
We experiment our frameworks on several grid-based navigation multi-goal to validate the effectiveness of the approach.
The results show that our added Optimal transport-based rewards are able to speed up the learning process of agents and outperforms several baselines on multi-task learning.

%% file: introduction.tex
\section{Introduction}

Recent advances in Reinforcement learning (RL) have achieved significant achievements in the field of gaming and robotics controls \cite{mnih2015human}, \cite{espeholt2018impala}. However, applying RL in practice is difficult since RL algorithms require a huge amount of data. Multi-task RL is a potential approach to improve data efficiency of RL algorithms \cite{d2020sharing}. In the settings of multi-task RL, multiple different but related RL problems are solved, simultaneously or consequently, by a single agent or several agents. By allowing the agents can share its experiences, it is expected that each agent can explore their environment better by utilizing knowledge from other agents, thus improving the data efficiency. 

Nevertheless, it is usually observed that multi-task RL poses additional challenges such as the dominance of behavior from one task to the other, or the negative interference of gradient between tasks, which makes the training of the multi-task RL unstable compared to when training separately \cite{teh2017distral, rusu2015policy}. Several lines of works have arised to mitigate the challenges in multi-task learning RL includes multi-objective optimization \cite{yu2020gradient, kurin2022defense}, incorporating additional domain-knowledge for solving tasks \cite{andrychowicz2017hindsight, sodhani2021multi}, leveraging past experience \cite{ yu2021conservative, singh2020cog}, discovering latent structure of tasks \cite{zhang2020learning, shu2017hierarchical} and policy regularization \cite{teh2017distral}. In this paper, we put our focus on the last approach.

Traditional works on knowledge distillation in RL use an information theory approach to inject behavior in behavior policy to the target task policy \cite{wu2019behavior,teh2017distral}. In this framework, a prior policy represents the common knowledge about the environment, and the learned policy is constrained to stay close to this prior policy, which results in a biased exploration of the learned policies toward potential state regions according to prior knowledge. In multi-task RL, the prior policy can be learned to capture a variety of common skills that are helpful across tasks. Task specific policies are trained to optimize the expected returns from its own environments while minimizing the deviation from the distill policy in terms of Kullback Leibler (KL) divergence. 

In this paper, we explore the use of Optimal transport (OT) in the context of knowledge distillation multi-task RL. Instead of minimizing an information regularizer, we directly constrain the state-action distributions of the prior and tasks policies by the Sinkhorn distances \cite{cuturi2013sinkhorn}. Unlike KL divergence, OT is a true metric that operates on a metric space, this makes the Optimal transport geometric-aware distance.
OT has gained traction in the machine learning community since the work of \cite{wgan} in GAN literature by solving a dual formulation of Wasserstein distance (WD).
In our work, we use Sinkhorn distance to approximate the WD.
The estimated Sinkhorn distance is used to construct an amortized reward, which is added to the standard reward from the environment.
We test the effects of our proposed framework on several grid-based environments with different multi-task state distribution structures.
The results show that our work can outperform the baseline SAC agents and is comparable with Distral  framework \cite{teh2017distral}.


Our paper is organized as follows: in Section II and III, the related works and background are presented; our proposed method is introduced in Section IV; the experiments are given in Section V; and Section VI is the conclusion.






%% file: related_work.tex
\section{Related Works}

In a single-agent/single-task scenario, the agent uses the reward from the environment to update its model parameters. Deep RL approaches employ deep networks to parameterize these mapping. For training the agent's model parameters, RL agents can use the experience collected by the agent itself in an on-policy training manner or it can use the experience from a different behavior policy in an off-policy training manner. Popular on-policy approaches include REINFORCE \cite{sutton2018reinforcement}, TRPO \cite{schulman2015trust}, PPO \cite{schulman2017proximal}. The on-policy methods usually have high variance in the estimated value functions and the requirement to use freshly sampled interactions every update step, which are not suitable in the settings of learning multi-task RL. In contrast, off-policy methods can overcome these issues by using bootstrapping mechanisms to make the off-policy value estimation biased. Popular off-policy approaches include DQN \cite{mnih2015human}  and SAC \cite{haarnoja2018soft}.

In multi-task RL settings, agents operate in separated environments with the same dynamics.
Popular frameworks employ settings in which each task has different reward functions.
As the dynamics of the environment are the same, agents can use the samples collected from a different learning task to update its parameters.
From the off-policy perspective, samples from the other learning tasks could be viewed as they are generated from behavior policies.
To correct the off-policy factor of the samples, IMPALA \cite{espeholt2018impala} introduces the V-trace term.
Similar works related to this direction include $Q(\lambda)$ \cite{harutyunyan2016q} and Retrace \cite{munos2016safe} algorithms.

Going to the knowledge distillation, DisTral \cite{teh2017distral} introduces a method for distilling knowledge between tasks by using KL regularization RL. In \cite{vuong2019sharing} proposed using an additional agent to decide the shareability of states. Other works transfer knowledge of tasks in the past by using reward shaping \cite{ng1999policy}, importance sampling \cite{tirinzoni2018importance} and offline RL \cite{singh2020cog}.


One of the key issues in co-training multi-task agents is to identify the similarity between two agent's policies. KL divergence is one of the most popular frameworks to evaluate such differences \cite{fox2015taming} due to its simplicity and applicability to existing algorithms. Recently, Optimal Transport has been frequently used as an alternative means to measure the discrepancy between policies in reinforcement learning. Several successful applications of OT such as in unsupervised RL \cite{he2022wasserstein} and Imitation learning \cite{dadashi2020primal, papagiannis2020imitation}.

%% file: background.tex
\section{Background}

\subsection{Preliminaries and Notations}
In this paper, we use the standard usual notation of MDP for a single task, denote by a 4-tuple of $(X, A, p, r)$. 
More specifically, an agent interacts with an environment by selecting an action in action space \(a \in A\) from observation \(x \in X\).
The agent policy is denoted as \(\pi(a|x)\).

Given the agent's policy, the next state observation \(x’ \in X\) is determined by a conditional probability on current state and action taken \(p(x’|x, a)\) and is dependent on the nature of the MDP. At every time step, the agent receives a reward signal \(r(x, a) \in \R\). We denote a trajectory \(\tau_\pi\) is a complete interaction sequence induced by following the policy \(\pi\) in the environment, \(\tau_\pi=(x_0,a_0,r_0,...)_\pi\). The standard RL objective is to find a policy that maximizes the total expected discounted reward over trajectories, with a discounted factor \(\gamma \in (0,1)\) that trades off between immediate and delayed rewards.
\begin{equation}
    J(\pi)=\E_{\pi}\Big[ \sum_{t} \gamma^t r(x_t, a_t) \Big] \label{eq:rl}
\end{equation}

In the multi-task context, we consider the setting that each task $i^{th}$, numbered from $1 \dots n$ with $n$ is the number of tasks, shares the same observation space $X$, action space $A$ and the transition probability $p$. Each task $i^{th}$, however, is equipped with a different reward function $r_i(x, a)$. Given each task $i^{th}$, the learned policy $\pi_i$ is trained to maximize the expected rewards for the task itself:
\begin{equation}
    J(\pi_i)=\E_{\pi_i}\Big[ \sum_{t} \gamma^t r_i(x_t, a_t) \Big] \label{eq:multirl1task}
\end{equation}
The RL objective from (\ref{eq:rl}) extends to capture the sum of total expected rewards from all the tasks as
\begin{equation}
    J(\pi_1, \pi_2, \dots, \pi_n) = \sum_{i = 1}^n J(\pi_i)
\end{equation}
The agents $i^{th}$ are co-trained simultaneously to maximize the total expected rewards $J(\pi_1, \pi_2, \dots, \pi_n) $

\subsection{KL regularized RL objective }

One of the traditional method for solving (\ref{eq:rl}) is to use Q learning \cite{watkins1992q}, which directly estimates the optimal value function Q. The exploratory policy is usually a stochastic policy that depends on the optimal policy induced by the estimated optimal value function,  which is typically an \(\varepsilon\)-greedy policy or noise-added estimated optimal policy \cite{fujimoto2018addressing}. The use of the min (or max) operator in the estimated optimal Q function introduces several problems: early commitment and overestimation \cite{hasselt2010double, fujimoto2018addressing}. Information regularized RL has been developed as a means to counteract these issues by avoiding using a deterministic decision rule of the max operator in Q learning and replacing it with a soft-max, the information constraint is conventionally a KL divergence. In the KL regularized RL framework \cite{fox2015taming}, we have the following objective.
\begin{equation}\label{eq:klrg}
    J_{\text{KL}}(\pi, \pi_0) = \E_\pi \Big[\sum_t \gamma^t r_t -\alpha_{\text{KL}} \gamma^t \text{KL}[\pi(\cdot | x_t)\| \pi_0 (\cdot | x_t)] \Big]
\end{equation}
Where \(\pi_0\) is called default or prior policy, this default policy plays the role as a means to exert knowledge into the RL agent in a sense that, when the agent’s behavior is not driven by reward signal, it should follow a default behavior. The default policy can be fixed prior to \cite{fox2015taming, haarnoja2018soft} or learned \cite{teh2017distral, galashov2019information} during training time. When nothing is known in the environment, this default policy can be set to be a uniform distribution over actions across all states, in which case the KL constraint in (\ref{eq:klrg}) returns to the traditional entropy regularized RL objective up to some constant factor \cite{galashov2019information, haarnoja2018soft}. 

\subsection{Knowledge distillation with KL regularized RL}

In the multi-task reinforcement learning problem, where multiple agents interact with the environment, each of which tries to solve their own respective tasks, common knowledge that is consistent in multiple tasks can be shared between agents to speed up the exploration process through the default policy. We start from Distral \cite{teh2017distral} that transfers knowledge in multi-task RL based on the KL regularization approach, a shared policy is trained to guide the task specific policies through KL regularization term. The Distral framework optimizes the objective function 
\begin{equation}
    J_{\text{Distral}} = \sum_{i=1}^n J_{\text{KL}}(\pi_i, \pi_0).
\end{equation}
Distral introduces a default policy, which acts as a central policy that keeps the common skills between all of the sharing tasks. The common skills are then used by each task policy to refine and fine-tune for their own specific goals. By using the KL divergence, behaviors that are outside the support of the default policy are generally discouraged since they make the KL tend to infinity. Throughout the training process, the learned default policy will expand its supported region to encompass all the task specific behaviors; the default policy is more or less the union of the task policies, which plays a role as a communication channel to transfer behaviors between multiple tasks.

%% file: method.tex
\section{Distill Knowledge with Optimal-Transport regularization}
We consider the problem of training multiple task policies, each of which optimizes their own total expected rewards with a regularization term. The regularization term is built upon the Optimal transport distance of state-action distribution of each agents.

To make the calculation tractable, we employ the Sinkhorn distance, which estimates the entropy regularized Wasserstein distance on the batches of tuples \((x_t, a_t)_{\pi}\) from trajectories \(\tau_\pi\) induced by each agent. The computed discrepancy is then used as amortized rewards derived to minimize the Optimal transport distances.




\subsection{Reward proxy from Sinkhorn distance}

Given the default policy \(\pi_0\) and a task policy \(\pi\), we study the problem of computing the reward proxy for distilling behaviors from \(\pi_0\) to \(\pi\).
The Sinkhorn distance \cite{cuturi2013sinkhorn} between two discrete distributions \(\rho_\pi\) and \(\rho_{\pi_0}\) of state-action by following \(\pi\) and \(\pi_{0}\) is
\begin{align}
     W(\rho_{\pi}, \rho_{\pi_0})&=\min_{P \in \mathcal{U}(\rho_\pi, \rho_{\pi_0})} \langle P, M\rangle + \lambda H(P)
    \label{eq:WD}
\end{align}
With \( \langle \cdot, \cdot  \rangle \) denotes the Frobenius dot-product and \(P\) belongs to the set \(\mathcal{U}(\rho_\pi, \rho_{\pi_0})\) of all non-negative matrices whose row and column sums are distributions of \(\rho_\pi\) and \(\rho_{\pi_0}\) respectively, \(H(M)\) is the entropy of the joint distribution \(P\) and \(\lambda\) is the Lagrangian dual variable. \(M\) is the cost matrix, \(M_{ij}\) denotes the distance \(d\) between the state-action pair of the \(i^{th}\) atom from \(\rho_\pi\) to the \(j^{th}\) atom from  \(\rho_{\pi_0}\). Given the optimal joint distribution \(P^*\), the contribution of each of the atom in each distribution to the total Sinkhorn distance can be calculated as 
\begin{equation}
\begin{array}{ll@{}ll}
c_\pi := (P^* \odot M)\textbf{1} &  c_{\pi_0} := (P^* \odot M)^T\textbf{1} \label{eq:contrib}
\end{array}
\end{equation}
where \(\textbf{1}\) is a vector of ones with appropriate dimension and \(\odot\) is element-wise product. From these vectors of contributions of each atom (which in this case is a state-action vector), we can evenly assign the (proxy) rewards to the state-action pairs based on how significant they contribute to the final OT distance, in optimizing this new reward, RL agents will effectively minimize the OT distance \cite{papagiannis2020imitation}. For clarity, each component of the contribution vector can be equivalently written out in full as
\begin{dmath}
    c_\pi((x, a)_\pi) =
    \sum_{(x, a)_{\pi_0} \in \tau_{\pi_0}} d\Big( (x, a)_\pi, (x, a)_{\pi_0} \Big) P^*\Big( (x, a)_\pi, (x, a)_{\pi_0} \Big)\label{eq:cont}
\end{dmath}
with \(d\) is the usual Euclidean norm. In this work, we restrict the distributions of \(\rho_\pi\) and \(\rho_{\pi_0}\) to be samples \((x, a)\) from single trajectories of the agent policy \(\tau_\pi\) and default policy \(\tau_{\pi_0}\). The distribution of \(\rho_\pi\) and \(\rho_{\pi_0}\) is approximated by the empirical distribution of data as the sum of Dirac functions centered at each data point.

The range of the Optimal transport distance is unbounded and depends heavily on the nature of the metric space being used, for example, normalized and unnormalized vector inputs will produce OT distances at different scales. Thus, to obtain reliable distance signals across state configurations that do not interfere too much with the actual reward signals in an uncontrollable way, it is necessary to map the calculated OT reward to a bounded range.

Inspired from \cite{dadashi2020primal}, the proxy reward of a state-action pair can be defined as a monotonically decreasing function of the contribution of that pair to the OT distance:
\begin{equation}
    s_{\pi} = \sigma \exp \left( \frac{\beta T}{\sqrt{|X| + |A|}}c_\pi \right) \label{eq:rw}
\end{equation}
Where \(c_\pi\) is calculated as in Eq. \ref{eq:cont}, to balance the influence of each dimension in the state-action, we normalize the state-actions to zero-mean and std one before calculating Euclidean distance \(d\). We use two hyper-parameters \(\sigma\) and \(\beta\), the \textit{scaling factor} and \textit{decaying factor} respectively, to control the proxy rewards. By the formula (\(\ref{eq:rw}\)), the bonus rewards are bounded between \((0, \sigma)\), the term \(\frac{T}{\sqrt{|X| + |A|}}\) normalizes the contribution \(c\) by dimensionality
and time horizon, and \(\beta\) is a coefficient term that controls the rate at which the rewards decay over distances. The calculated proxy reward in Eq. \ref{eq:rw} is then added with the original reward from the tasks to form a final reward function.

\subsection{Distillation between multiple agents}
The task reward is augmented with the proxy reward calculated in the previous section, providing smooth, transient reward signals to guide task agents  to specific behaviors that are frequently exploited by agents of other tasks. The details for our algorithm is described in Algorithm \ref{al:a2}.

In the proposed sharing framework, we employ the settings that each agent is trained simultaneously. More specifically, at each training batch, the agent $i$ is let to interact with the environment to collect the rollout trajectories. The reward of every timestep from the collected trajectories is then augmented with the proxy reward in eq.($\ref{eq:contrib}$). 
The newly obtained rewards are then treated as usual rewards as in traditional RL problems and can be optimized by RL algorithms. In this work, we choose to use SAC \cite{haarnoja2018soft} as the optimization RL algorithm.



\begin{algorithm}
\caption{Distill OT Algorithm based on SAC}\label{al:a2}
\hspace*{\algorithmicindent} \textbf{Input:} \par
\hskip\algorithmicindent \text{ $n$: number of tasks, \(\sigma\): scaling factor, $\beta$: decaying factor, } \par
        \hskip\algorithmicindent \text{ $B_i$ and $\pi_i$: replay buffer and the policy of $i^{th}$ task} \par
        
\begin{algorithmic}[1]

\While{not converge}
    \State $\tau \text{   := \{\}}$
    \For {$i=1..n$}
        \State \text{$\tau_i$ := $\{x_0, a_0, r_0, ...\}$, an episode with $\pi_i$ }
        \State \text{Add $\tau_i$ to $\tau$}
    \EndFor
    \For{$i=1..n$}
        \State \text{Let $j$ be a random task,  $j \neq i$}
        \State \text{Let $\tau_i$ be the rollout of the task $i$ from $\tau$}
        \State \text{Let $\tau_j$ be the rollout of the task $j$ from $\tau$}
        \State \text{$s^t_i$ := compute\_reward($\tau_i$, $\tau_j$) according to eq.($\ref{eq:contrib}$) } \par
        \hskip\algorithmicindent \text{ and eq.($\ref{eq:rw}$) for each timestep $t$}
        \State $r^t_i = r^t_i + s^t_i$
        \State \text{Add $\tau_i$ to buffer $B_i$}
        \State \text{Update $\pi_i$ by SAC}
    \EndFor
\EndWhile

\end{algorithmic}
\end{algorithm}

In order to compute the OT distances, one needs samples from both distributions. This makes the approach of using the shared policy less appealing, as we need to allocate environment interactions to an additional sharing policy \(\pi_0\), which will be discarded after the training.
To bypass the need of this sharing policy, we propose an approach to circumvent this issue by directly comparing distances between random pairs of trajectories (line 8-10).
There are several alternatives, such as aggregating all task trajectories into a large trajectory that represent the distribution of all tasks. However, we empirically find that the random-pair approach is simple and effective enough in practice.

From the Eq. $\ref{eq:rw}$, maximizing $s_i$ will lead to reducing the OT distance from two trajectories $\tau_i$ and $\tau_j$.
In the training process of policy $\pi_i$ of the agent $i$, iteratively  choosing $j$ randomly results in distilling common knowledge from all of the tasks into the specific policy $\pi_i$ of task $i$.
The scaling parameter $\sigma$ can be chosen to balance the influence of proxy rewards on the environment rewards.

The number of roll-outs per training step, number of random selected tasks $\tau_j$ can be considered as hyper-parameters.


%% file: result.tex
\section{Experiments and Results}
\subsection{Experiments on grid world environments}
\begin{figure}[th]
\begin{subfigure}{.16\textwidth}
  \centering
  \includegraphics[width=.96\linewidth]{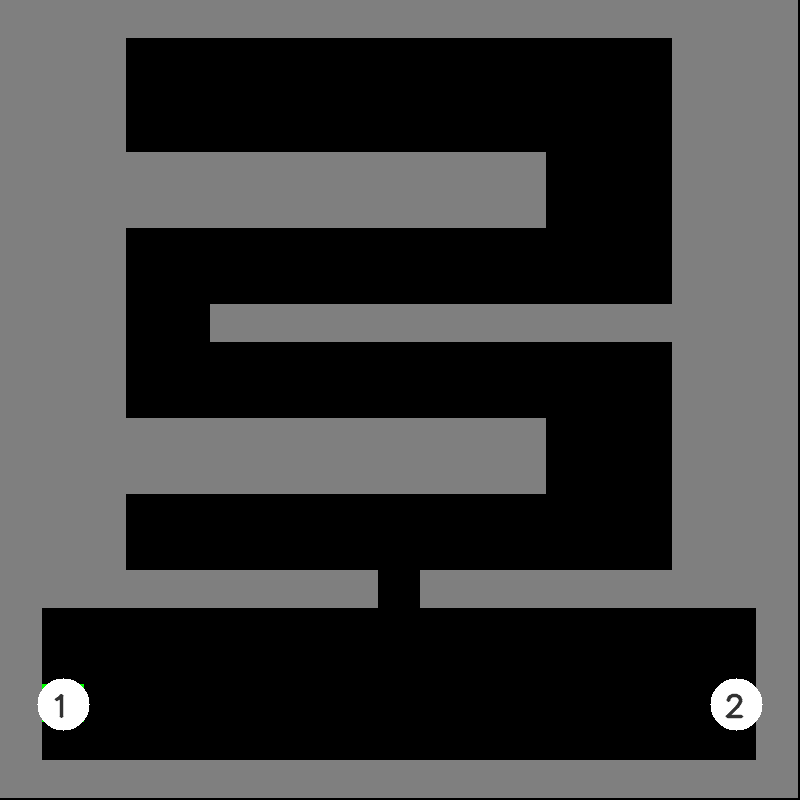}
  \caption{\textit{Zig-zag corridor}}
  \label{fig:sfig3}
\end{subfigure}%
\begin{subfigure}{.16\textwidth}
  \centering
  \includegraphics[width=.96\linewidth]{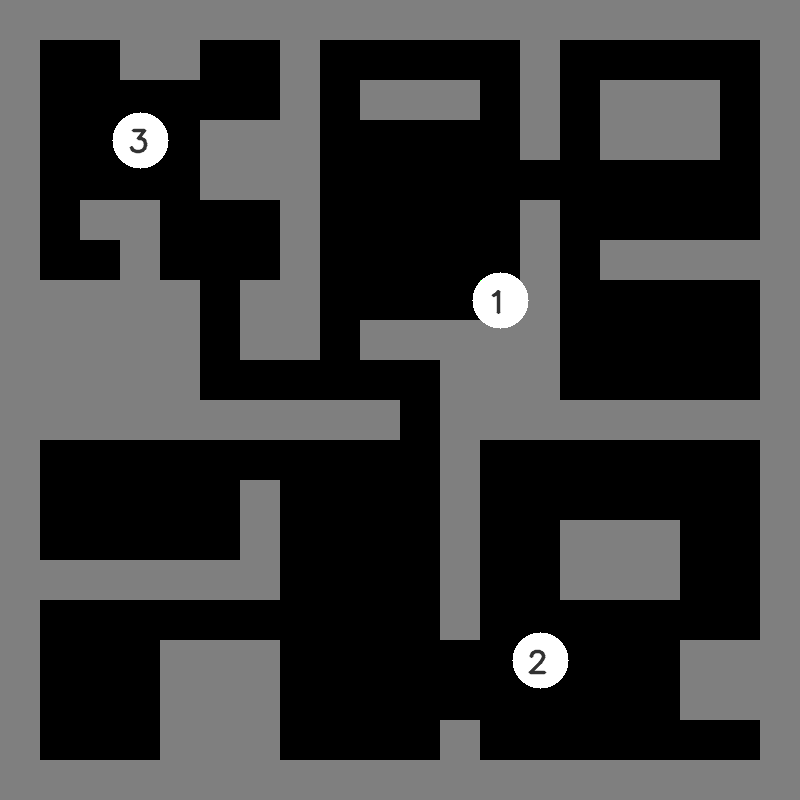}
  \caption{\textit{Maze}}
  \label{fig:sfig2}
\end{subfigure}%
\begin{subfigure}{.16\textwidth}
  \centering
  \includegraphics[width=.96\linewidth]{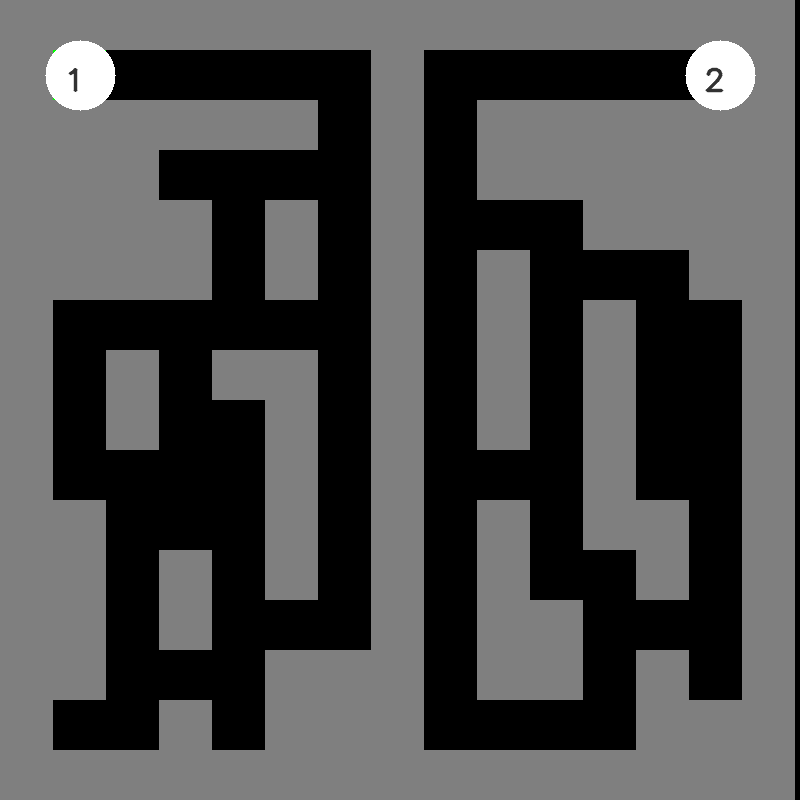}
  \caption{\textit{Separated mazes}}
  \label{fig:sfig1}
\end{subfigure}%

\caption{The map of three tested environments with its multi-task goals, numbered by 1, 2, 3 in each specific figures.}
\label{fig:gridworld}
\end{figure}

We choose grid world-based environments with different wall layouts to validate our effectiveness of the proposed knowledge sharing method. In general, a specific agent has to navigate to a target location through four possible actions of moving left, right, up and down. Each task is differ by the target location.
The specific layout of environment and task goals are illustrated in Fig \ref{fig:gridworld}.
The observations of the agents are the normalized coordinate of its current location in the grid world.
For each task, the starting states of agents are uniformly distributed across all the maps.
Agent receives a small penalty every time step or when hitting walls and a positive reward upon reaching the goal, the final positive reward is scaled appropriately with the size of the grid map so that difficult tasks get more rewards when successful.
The reward scheme is designed to encourage the agents to find the shortest possible path to the goals while avoiding hitting walls.
The specific details of three environments are described as follows:

\begin{itemize}
    \item \textit{Zig-zag Corridor} (Fig. \ref{fig:sfig3}): All agents have to pass through a long, zig-zag corridor to get into a room in which each goal of tasks is located at different ends. The long corridor is the state region that should be shared between tasks and it occupies a large portion of the environment. Therefore, most of the knowledge can be shared among the two agents.
    
    \item \textit{Maze} (Fig. \ref{fig:sfig2}): Behaviors of every agent can be quite different since each task requires the agent to go to a separate part of the maze. In this environment, as behaviors of some tasks are conflicting with each other, sharing algorithms should have some mechanisms to balance out the influences of the knowledge from other tasks and the performance of current tasks. Only a part of knowledge can be shared in this environment.

    \item \textit{Separated Mazes} (Fig. \ref{fig:sfig1}): Two parts of the maze are separated from each other,  where each goal lies in its respective part. Agents randomly spawn in each part. If the agent of one task is spawned at the other task's maze, it can not move to its goal and thus the best actions are to avoid hitting walls. This environment is designed so that the distilled knowledge between tasks is minimal.

\end{itemize}

In each experiment, we compare our method \textbf{OT-sharing} with the results from \textbf{Distral} \cite{teh2017distral} and the settings where each task is trained separately, \textbf{No-sharing}. The network structures are kept the same for all experiments with three layers of fully connected network with 256 hidden units each layer and non-linear \textit{Tanh} activation functions for both value functions and policies. Network’s weights are updated using Adam Optimizer. We repeat all experiments with 6 different seeds and average the results over all seeds. The details of the hyper-parameters being used are described in Table \ref{table:hyper}.
\begin{table}[htbp]
\centering
\caption{Hyper-parameter settings}
\label{table:hyper} 
\catcode`,=\active
\def,{\char`,\allowbreak}
\renewcommand\arraystretch{1.2}
\begin{tabular}{p{4cm}<{\raggedright} p{3.5cm} }
  \toprule
    \textbf{Hyperparameters}             & \textbf{Values}           \\ 
  \midrule
    Num.seeds                   & 6 \\
    Num.timesteps               & 200k                      \\
    Buffer size                 & 10k                      \\
    Horizon (\(T\))             & 100                     \\
    Minibatch size              & 128                       \\
    Discount (\(\gamma\))       & 0.99                      \\
    Adam stepsize               & \(10^{-3}\)     \\ 
    Polyak                       &  0.99\\
    $\alpha$ (Distral)                & 0.5\\
    $\beta$ (Distral)                & 5\\
    alpha SAC                   & 0.1\\
    $\beta$ (eq. \ref{eq:rw})      & 2, 5\\
    $\sigma$ (eq. \ref{eq:rw})    & 0.1\\
  \bottomrule
\end{tabular} 
\end{table}


\begin{figure}
\begin{minipage}{.5\linewidth}
\centering
\subfloat[\textit{Zig-zag Corridor}]{\label{fig:rand2}\includegraphics[width=4.3cm]{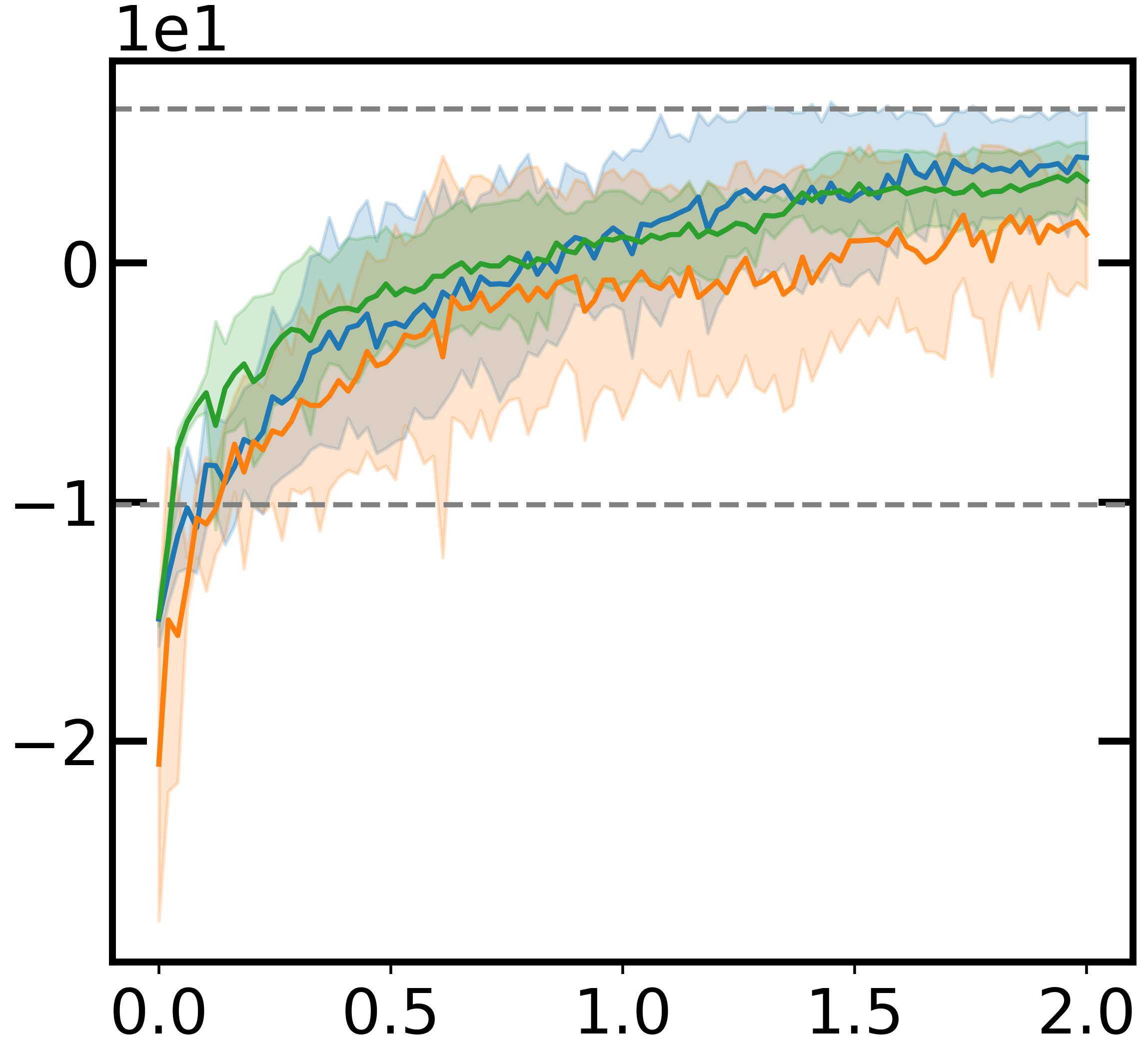}}
\end{minipage}%
\begin{minipage}{.6\linewidth}
\centering
\subfloat[\textit{Maze}]{\label{fig:rand4}\includegraphics[width=4.3cm]{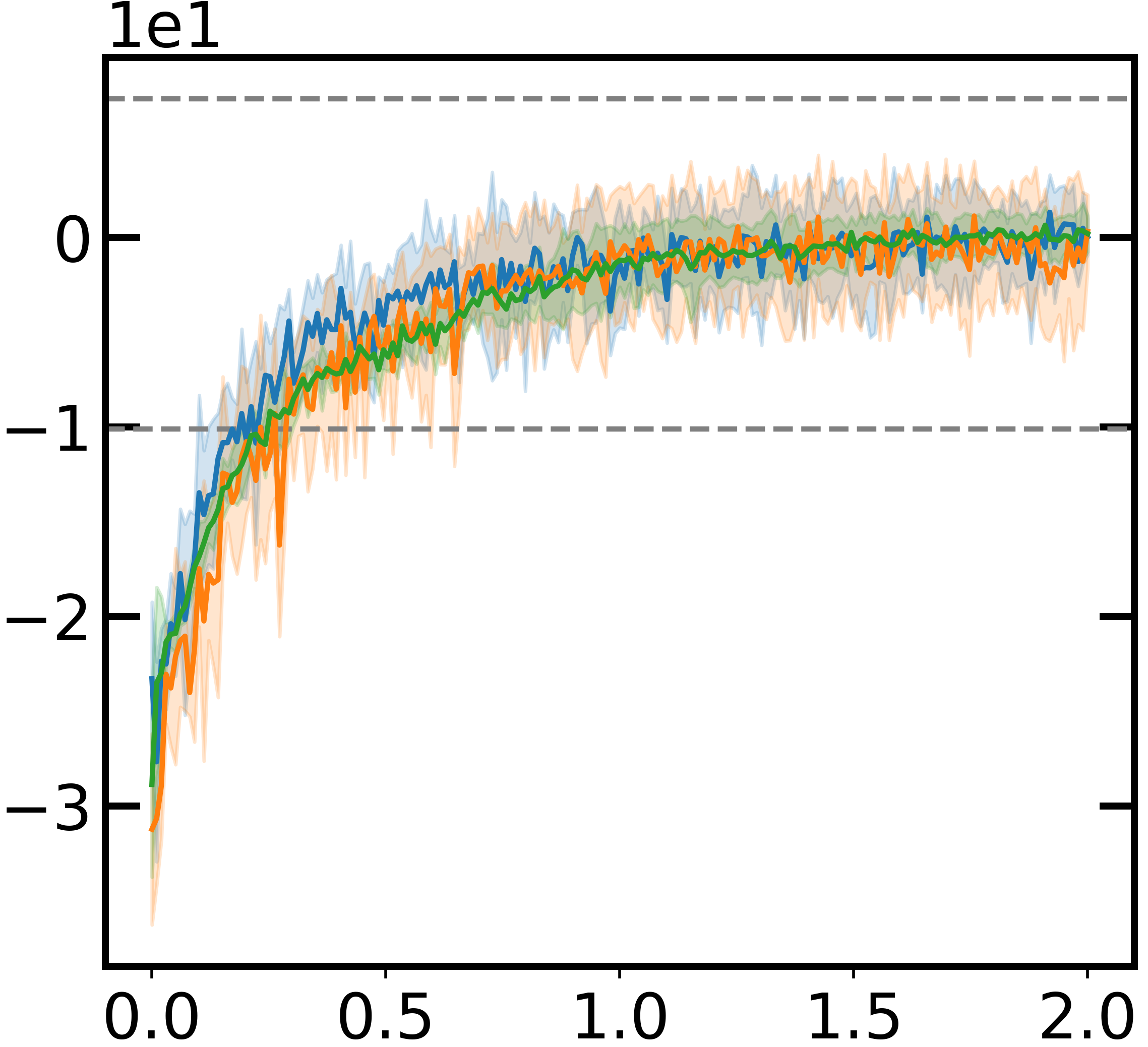}}
\end{minipage}\par\medskip
\centering
\subfloat[\textit{Separated Mazes}]{\label{fig:rand6}\includegraphics[width=6.5cm]{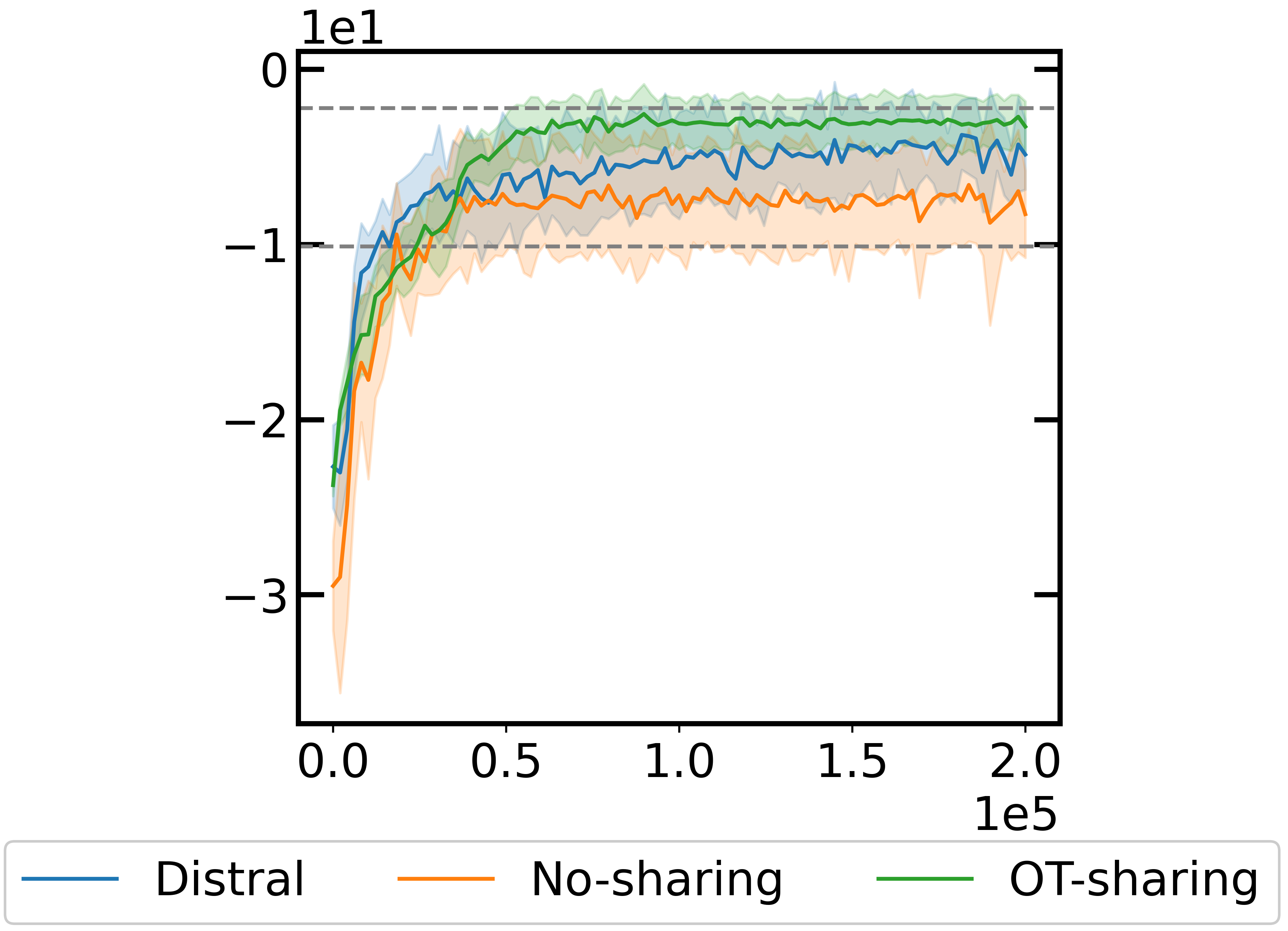}}
\caption{Training curves on different environments, averaged over 6 seeds in each experiment. The upper horizontal dash lines are the expected return of the optimal policies, while the lower ones denote a benchmark  performance where agents do not hit walls.}
\label{fig:res}
\end{figure}

Results from the three experiments are shown in the Table \ref{tab:res}, with the mean and std of total returns averaged from all experiments. Our method achieves a comparable performance with \textbf{Distral} and \textbf{No-sharing} in the \textit{Maze} environment, and surpasses both in \textit{Separated Mazes}.
It shows that our proposed method can achieve more stable results when the amount of distill knowledge is minimal.
We observe a correlation of sharing agents’ performance between \textit{Zig-zag} and \textit{Separated Mazes}, as both sharing methods significantly boost the performance of all agents, while in \textit{Maze} all algorithms behave almost similarly.


The training curves are shown in the Fig. \ref{fig:res} that illustrates the asymptotic behaviors of each algorithm.
The lower dash line at -10 marks the point when the agents learn to not hit the wall.
The upper dash line is the expected optimal return.
Compared to the baseline of \textbf{No-sharing}, our method exhibits a better convergence rate which learns faster in \textit{Zig-zag} and \textit{Separated Mazes} and is comparable in the \textit{Maze} environment. With \textbf{Distral}, our convergences are equivalent in two environments \textit{Zig-zag} and \textit{Maze} and surpass in \textit{Separated Mazes}.

\begin{center}

\begin{table}
\centering
\begin{tabular}{ |c|c|c|c|c|c| } 
\hline
\multicolumn{2}{|c|}{Task} & \textbf{No-share} & \textbf{Distral} & \textbf{OT-sharing} & Opt\\
\hline
\parbox[0]{10mm}{\multirow{3}{*}{{\textbf{Zig-zag}}}}
&  1 &  3.7$\pm$3.7 & 4.5$\pm$2.0 & 2.9$\pm$3.1 & 6.4  \\ 
&  2 & -0.7$\pm$2.9 &  4.4$\pm$1.9 & 4.2$\pm$1.5 & 6.4 \\ 
& avg & 1.5$\pm$2.2 & \textbf{4.4$\pm$1.7} &  3.6$\pm$1.4 & 6.4  \\
\hline
\parbox[l]{10mm}{\multirow{4}{*}{{\textbf{Maze}}}}
&  1 &  -3.7$\pm$5.6 & -3.0$\pm$5.4 & -3.5$\pm$1.0 & 7.5  \\ 
&  2 &  5.1$\pm$4.5 &  6.6$\pm$1.2 & 5.5$\pm$1.2 &  7.5  \\ 
&  3 & -3.6$\pm$6.5 & -3.7$\pm$3.6 & -1.4$\pm$3.8 & 6.9 \\
& avg & -0.7$\pm$3.4 & \textbf{0.1$\pm$1.7} & \textbf{ 0.1$\pm$1.1} & 7.3 \\
\hline
\parbox[]{13mm}{\multirow{3}{*}{{\textbf{Separated}}}}
&  1 & -8.0$\pm$4.0 & -5.2$\pm$3.8 & -3.5$\pm$2.5 &  -2.2 \\ 
&  2 &  -7.2$\pm$4.7 & -3.9$\pm$3.5 & -2.4$\pm$0.7 & -2.2 \\ 
& avg & -7.6$\pm$2.8 & -4.6$\pm$2.2 & \textbf{-3.0$\pm$1.3} & -2.2  \\
\hline
\end{tabular}
\caption{\label{tab:res}The average return and standard deviation of all experiments over the last 4000  environment steps for each task. Opt is the performance of the optimal agents.}
\end{table}%

\end{center}

\subsection{Effects of Distill Knowledge}

\begin{figure}
    \centering
    \makebox[0pt]{
    \includegraphics[width=.52\textwidth]{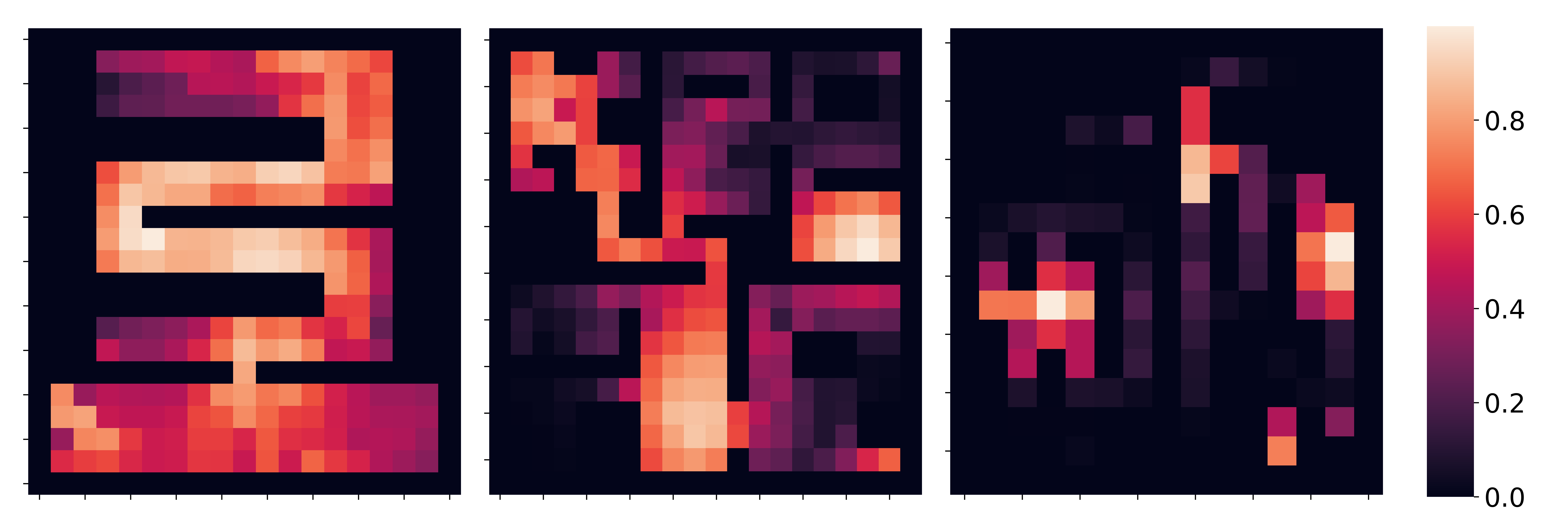}}
    \caption{Proxy reward calculated from Sinkhorn distances in three environments. From left to right: Zig-zag corridor, Maze and Separated mazes.}
    \label{fig:reward}
\end{figure}

In this section, we conduct additional experiments to analyze the effects of the OT rewards on all environments. The augmented rewards are averaged over all tasks in each environment and are shown in Fig \ref{fig:reward}. In this experiment, the optimal transport rewards for all states in the gridworld are calculated by Eq. \ref{eq:rw}, each corresponding to a certain degree to which task agents visit that state. Particularly, the higher the OT reward of a state is, the more likely that the state will be visited by some task policy. 

In the \textit{Zig-zag} environment, the OT rewards are distributed densely inside the Corridor, meaning that agents travel through this part frequently. The knowledge can be transferred easily from one agent to another. Similarly, in the \textit{Maze}, several spots with high OT rewards are visible that roughly coincide with the goal positions of some tasks. In the extreme case of the \textit{Separated Mazes}, the OT rewards are distributed randomly over the two separated parts.
It can be explained that the agents move randomly in this part until episodes end when they can not  reach their goals.